%% file: main.tex
\documentclass[10pt]{article} 

\usepackage[accepted]{rlj}           
\usepackage{bm}
\usepackage{algorithm}
\usepackage{algpseudocode}
\algrenewcommand\algorithmicrequire{\textbf{Input:}}
\algrenewcommand\algorithmicensure{\textbf{Output:}}

\usepackage{amssymb}            
\usepackage{mathtools}          
\usepackage{mathrsfs}           
\usepackage{graphicx}           
\usepackage{subcaption}         
\usepackage[space]{grffile}     
\usepackage{url}                
\usepackage{lipsum}             
\usepackage{siunitx}            

\definecolor{revcolor}{RGB}{146,32,186}

\hypersetup{
  pdftitle={Dynamics Models for Offline Hyperparameter Selection in Real-World RL},
  pdfauthor={Jordan Coblin, Han Wang, Martha White, Adam White},
  pdfkeywords={Applied RL, Dynamics Models, Hyperparameter Selection, Water Treatment, Industrial Control},
}


\title{Dynamics Models for Offline Hyperparameter \\
Selection in Real-World RL}

\setrunningtitle{Dynamics Models for Offline Hyperparameter Selection in Real-World RL}

\author{
  Jordan Coblin\textsuperscript{1}, Han Wang\textsuperscript{1}, Martha White\textsuperscript{1,2,3}, Adam White\textsuperscript{1,2,3}
}

\emails{\{coblin,han8,whitem,amw8\}@ualberta.ca}

\affiliations{
$^{1}$\textbf{University of Alberta} \\
$^{2}$\textbf{Alberta Machine Intelligence Institute (Amii)} \\
$^{3}$\textbf{Canada CIFAR AI Chair at Amii}
}

\contribution{
  We present the first application of calibration models for offline hyperparameter selection in a real-world industrial setting, focusing on sensor prediction tasks from a municipal water treatment plant.
}
{
  Calibration models for offline hyperparameter selection were introduced by \citet{wang2022pesky}, but were evaluated only in simple simulated domains. Although \citet{janjua2023wtprediction} studied prediction tasks in a water treatment plant, that work did not consider calibration models for hyperparameter selection.
}

\contribution{
  We extend empirical evaluation methods for calibration models by introducing analyses of rollout quality, hyperparameter sensitivity, and the use of dynamic time warping to assess alignment between model-generated and true trajectories. We use these methods to compare the performance of several calibration model architectures.
}
{
  Previous work has focused primarily on comparing performance of the best hyperparameter configuration, but has not examined measures of model accuracy or hyperparameter sensitivity.
}

\contribution{
  We bridge the gap toward real-world deployment by scaling calibration models to a year-long offline dataset and investigating their ability to simulate distribution shifts for the fine-tuning setting.
}
{
  Prior research has focused on small-scale, simulated domains and has not addressed the critical challenges of scalability and adaptation under realistic distribution shifts, both of which are essential for deploying reinforcement learning systems in practice.
}


\summary{
  A key obstacle to deploying reinforcement learning in real-world systems is hyperparameter selection, particularly when simulators are unavailable and online experimentation is costly. Prior work has proposed calibration models trained on offline data to approximate environment dynamics and enable offline hyperparameter selection, but these methods have so far been evaluated only in simple simulated settings. In this paper, we present the first application of calibration models in a real-world industrial setting: a municipal water treatment plant. We evaluate several calibration model approaches, including a $k$-nearest neighbors model with a Laplacian distance metric, on high-dimensional, non-stationary sensor data for nexting prediction tasks. Our results show that these models can generate realistic long-horizon rollouts and recover meaningful hyperparameter sensitivity trends. We further examine how calibration models scale to year-long datasets, how they support the selection of fine-tuning learning rates for pre-trained agents, and how robust they are under distribution shift. Overall, our findings provide a proof of concept for using offline dynamics models to support RL deployment in real-world environments, while highlighting important practical challenges for future work.
}

\begin{document}

\makeCover 
\maketitle  

\begin{abstract}

    A key obstacle to deploying reinforcement learning in real-world systems is hyperparameter selection, particularly when simulators are unavailable and online experimentation is costly. Prior work has proposed calibration models trained on offline data to approximate environment dynamics and enable offline hyperparameter selection, but these methods have so far been evaluated only in simple simulated settings. In this paper, we present the first application of calibration models in a real-world industrial setting: a municipal water treatment plant. We evaluate several calibration model approaches, including a $k$-nearest neighbors model with a Laplacian distance metric, on high-dimensional, non-stationary sensor data for nexting prediction tasks. Our results show that these models can generate realistic long-horizon rollouts and recover meaningful hyperparameter sensitivity trends. We further examine how calibration models scale to year-long datasets, how they support the selection of fine-tuning learning rates for pre-trained agents, and how robust they are under distribution shift. Overall, our findings provide a proof of concept for using offline dynamics models to support RL deployment in real-world environments, while highlighting important practical challenges for future work.

\end{abstract}

\input{sections/Introduction}

\input{sections/Background}

\input{sections/WaterTreatment}
\input{sections/RealWorldDeployment}
\input{sections/Conclusion}
\appendix








\bibliography{main}
\bibliographystyle{rlj}

\input{sections/Supplementary}


\end{document}

%% file: sections/Introduction.tex
\section{Introduction}
\label{sec:intro}

Reinforcement learning (RL) offers adaptive, data-driven control across a range of industrial applications, including assembly line automation \citep{tortorelli2022parallel}, thermal power generation \citep{zhan2022deepthermal}, and commercial cooling \citep{luo2022controlling}. However, deploying RL in real-world systems remains constrained by numerous practical challenges. Among these, hyperparameter selection has received comparatively little attention, despite RL performance being highly sensitive to hyperparameters such as learning rate, exploration schedule, and model architecture \citep{henderson2018rlmatters, andrychowicz2020matters, eimer2023hyperparameters}. There are several standard approaches to hyperparameter selection in real-world tasks, each with its own limitations. Default hyperparameters \citep{degrave2022magnetic} rarely reflect the dynamics of the target system and can leave substantial performance unrealized. Simulator-based tuning \citep{levine2016visuomotor, openai2019dexterous} requires an accurate simulator, which is often unavailable. Lastly, direct tuning in the real environment \citep{azuatalam2020hvac, luo2022controlling} is often infeasible when interactions are costly, risky, or time-intensive.

A promising alternative is to use the large offline datasets commonly available in industrial systems to select hyperparameters for agents that will subsequently learn online --- a setting known as Data2Online \citep{wang2022pesky}. This approach trains a dynamics model on offline logs and uses it as a surrogate environment, referred to as a \textit{calibration model}. Candidate agent configurations are evaluated through interaction with this model, and the best-performing configuration is selected for deployment and continued online learning in the real system. Unlike in standard model-based RL, the model is not used to optimize the deployment policy directly; its purpose is to preserve the relative performance of candidate hyperparameter configurations. Although prior work suggests that calibration models can approximate environment dynamics well enough for this purpose, they have so far been evaluated only in simple simulated domains, leaving their effectiveness in real-world systems unclear.

In this work, we extend the calibration model framework to sensor prediction tasks in a water treatment plant (WTP) in Drayton Valley, Alberta, Canada. Our contributions are threefold: (1) we present the first application of calibration models for offline hyperparameter selection in a real industrial system; (2) we extend evaluation methods of calibration models, incorporating rollout quality analysis, hyperparameter sensitivity, and dynamic time warping to assess trajectory similarity; and (3) we investigate scalability to large offline datasets and robustness under distribution shifts, addressing challenges critical to real-world RL deployment.

\paragraph{\texorpdfstring{Connections to time-series forecasting and model-based RL.}{Connections to time-series forecasting and model-based RL.}}
In the passive prediction setting studied here, calibration models resemble multivariate time-series forecasters: because the agent does not act on the plant, the model generates future sensor trajectories without conditioning on actions. This connects our setting to forecasting methods for multi-step sequence prediction \citep{salinas2020deepar, oreshkin2020nbeats, nie2023patchtst}. In control settings, by contrast, calibration models must represent action-conditioned dynamics, making them more closely related to learned dynamics models in model-based RL \citep{deisenroth2011pilco, chua2018Pets, janner2019mbpo}. The key distinction from both fields is the model's purpose. Rather than minimizing forecast error or directly optimizing a deployment policy, a calibration model serves as an evaluation environment for comparing hyperparameter configurations of an agent that will subsequently learn online. Its utility therefore depends on preserving hyperparameter rankings over long rollouts, which need not coincide with predictive accuracy.

%% file: sections/Background.tex
\section{Background}\label{sec:background}

This section introduces the framework and methods used for offline hyperparameter selection with calibration models. We first formalize the problem setting and calibration objective, then describe the k-nearest-neighbor ($k\text{NN}$) model used to approximate environment dynamics. We next introduce the nexting prediction task studied in our experiments and conclude with dynamic time warping, which we use to evaluate calibration-model rollouts.

\subsection{Problem Formulation}\label{subsec:problem_formulation}

Let $\mathcal{D} = \{(s_i, a_i, r_i, s_i')\}_{i=1}^N$ be a dataset of $N$ transitions sampled from a Markov decision process (MDP) under a behavior policy $\pi_\beta$, where $s_i \in \mathcal{S}$ is the state, $a_i \in \mathcal{A}$ is the action, $r_i \in \mathbb{R}$ is the reward, and $s_i' \in \mathcal{S}$ is the next state. A dynamics model $\hat{p}(s', r | s, a)$ is trained on $\mathcal{D}$ to approximate the environment's true transition function $p(s', r | s, a)$.

Following \citet{wang2022pesky}, we refer to $\hat{p}$ as a \textit{calibration model}: it is used solely to evaluate hyperparameter configurations for an agent that will subsequently learn online.

Let $\mathscr{A}$ be a learning algorithm, $\Lambda$ the hyperparameter space, and $\lambda \in \Lambda$ a hyperparameter configuration. For each $\lambda$, we define a sequence of policies $\{ \pi_t^\lambda \}_{t=0}^\infty$ as the result of running $\mathscr{A}$ interactively in the environment $p$, i.e.
$\{ \pi_t^\lambda \} = \mathscr{A}(\lambda; p)$. The expected return in the true environment is defined as:
\begin{equation*}
J_{\text{env}}(\lambda) = \mathbb{E}_{p} \left[ \sum_{t=0}^{\infty} \gamma^t r_t \,\Big|\, a_t \sim \pi_t^\lambda(\cdot | s_t),\, (s_{t+1}, r_t) \sim p(\cdot | s_t, a_t) \right],
\end{equation*}
where $\gamma \in [0, 1]$ is the discount factor. Similarly, the expected return using a calibration model is $J_{\text{model}}(\lambda) = \mathbb{E}_{\hat{p}} \left[ \sum_{t=0}^{\infty} \gamma^t r_t \,\Big|\, a_t \sim \pi_t^\lambda(\cdot | s_t),\, (s_{t+1}, r_t) \sim \hat{p}(\cdot | s_t, a_t) \right]$, where the policy sequence $\{ \pi_t^\lambda \}$ is now learned via interaction with $\hat{p}$, $\{ \pi_t^\lambda \} = \mathscr{A}(\lambda; \hat{p})$.

We define the optimal hyperparameters in each case as $\lambda_{\text{env}}^\star = \arg\max_{\lambda \in \Lambda} J_{\text{env}}(\lambda)$ and $\lambda_{\text{model}}^\star = \arg\max_{\lambda \in \Lambda} J_{\text{model}}(\lambda)$, and aim to learn a calibration model $\hat{p}$ such that $\lambda_{\text{model}}^\star = \lambda_{\text{env}}^\star$.

\subsection{\texorpdfstring{$\bm{k}\text{NN}$}{kNN} Calibration Model}\label{subsec:knn_calibration_model}

A good calibration model must be stable under long horizon rollouts, since hundreds or thousands of steps are typically necessary to evaluate a hyperparameter configuration. However, typical dynamics models are known to suffer from compounding errors, which can lead to significant divergence from the true environment over long horizons \citep{talvitie2017self,lambert2022compounding}. In order to mitigate this, \citet{wang2022pesky} propose using a non-parametric $k\text{NN}$ model to estimate $p$  by predicting next states and rewards using only transitions within the dataset, avoiding extrapolation into unobserved regions.


The $k\text{NN}$ model samples transitions from $\mathcal{D}$ to find the $k$ nearest neighbors of a given state-action pair according to some distance metric. Instead of using distance in the raw state-action space to find neighbors, an approximate Laplacian representation \citep{wu2018laplacian} can be used to construct a distance metric that is sensitive to the underlying structure of the MDP. We refer to a model that uses this distance metric as a \textit{Laplacian $k\text{NN}$} model, and a model that uses distance in the raw state-action space as a \textit{Euclidean $k\text{NN}$} model.

To address epistemic uncertainty, we use a \textit{leave-one-block-out} (LOBO) ensemble of five $k\text{NN}$ models. The dataset $\mathcal{D}$ is partitioned into five contiguous blocks, and each model is trained on a subset of $\mathcal{D}$ that excludes a different block. Results across models are aggregated by taking the \textit{worst rank} of a given hyperparameter configuration across all models, providing a conservative estimate of hyperparameter performance.

\subsection{Nexting Prediction}\label{subsec:nexting}

The \textit{nexting prediction} problem involves an agent predicting the discounted sum of future values of an observation signal, framed as a general value function (GVF) \citep{modayil2012nexting}. This formulation enables temporally extended predictions about scalar signals (often called \textit{cumulants}) in an environment. Following \citet{janjua2023wtprediction}, in this work we focus on the nexting prediction problem for sensors in a WTP\@. For a signal $o^i_t$ at time $t$, associated with channel or sensor $i$, the nexting value function is defined as:

\begin{equation}
v^i_t(s) = \mathbb{E} \left[ G^i_t \mid s_t = s \right], \quad G^i_t = \sum_{k=0}^\infty \gamma^k o^i_{t+k+1},
\end{equation}

where $\gamma \in [0, 1]$ is the discount factor controlling the timescale of the prediction. This task instantiates the formulation of Section~\ref{subsec:problem_formulation}: the cumulant $o^i$ plays the role of the reward, and the nexting target $G^i_t$ is the corresponding return. Because the agent's task is prediction rather than control, agent performance is measured by predictive accuracy rather than by achieved return, using the root mean squared error (RMSE) between the predicted value $\hat{v}^i_t$ and the empirical Monte Carlo return $G^i_t$ from data trajectories across all time steps $t \in \{0, 1, \ldots, T\}$. To account for differences in scale across sensors, we typically report the normalized RMSE (NRMSE), which divides the RMSE by the average return $\mu^i = \frac{1}{T+1} \sum_{t=0}^{T} G^i_t$. For hyperparameter selection, low prediction error thus takes the place of high return in the objective $J(\lambda)$.

    


\subsection{Dynamic Time Warping}

Dynamic time warping (DTW) is a technique for measuring similarity between time series that may differ in speed, timing, or phase \citep{kruskal1983dtw}. Given two sequences, DTW uses warping functions $\phi_x$ and $\phi_y$ to flexibly align elements along a common time axis, allowing sections to “stretch” or “compress” for better matching, even when sequences are out of sync or unevenly sampled. Alignment is guided by constraints collectively defined by a \textit{step pattern}; we use four common step patterns \citep{giorgino2009dtw} to mitigate sensitivity to any single choice.


%% file: sections/WaterTreatment.tex
\section{Real-World Application: Water Treatment Plant}
\label{sec:WaterTreatment}

To evaluate calibration models beyond simulation, we apply them to a working membrane-filtration pilot at a water treatment plant in Drayton Valley, Alberta, Canada. This setting presents high-dimensional, non-stationary, and noisy sensor data characteristic of real-world systems. Moreover, because no simulator is available and direct interaction is constrained by safety and data-collection considerations, calibration models provide a practical approach to offline evaluation.

\subsection{\texorpdfstring{Learning Task}{Learning Task}}

The task considered throughout this section is the nexting \emph{prediction} problem introduced in Section~\ref{subsec:nexting}. At each time step, the agent predicts the discounted sum of future values for each sensor signal over a horizon determined by $\gamma$. For example, the agent may predict the near-term evolution of membrane pressure. Such anticipatory predictions provide useful knowledge about the plant, supporting system monitoring and serving as a step toward closed-loop control \citep{modayil2012nexting, janjua2023wtprediction}. The agent is passive: sensor signals serve as prediction targets (cumulants), there is no extrinsic reward, and the agent takes no actions that influence the plant. Although prior work has considered WTP control variables such as chemical dosing rates and backwash schedules \citep{liu2022wtp}, we focus on prediction as an initial setting in which to establish the feasibility of calibration models.

    \subsection{Dataset}

    The offline dataset consists of over two years of sensor logs (collected between 2022 and 2024) from the WTP, with \(480\) sensor channels sampled at 1 Hz. For our initial experiments, we use a one-week slice of the dataset, which contains $\sim 3.5\times 10^{5}$ transitions. We discard sensors with consistently missing data and constant values, leaving us with a \(142\)-dimensional feature vector for our prediction agents and calibration models. In general, we leverage the data processing pipeline outlined in \citet{janjua2023wtprediction}.

    For our experiments, we consider three sensors that are both critical for control of the plant and provide variety in dynamics to predict. These are the membrane pressure (PIT300), influent temperature (TIT101), and influent turbidity (TUIT101) sensors.

    \subsection{Experimental Setup}

    We train four calibration models: a LOBO Euclidean $k\text{NN}$ ensemble, a LOBO Laplacian $k\text{NN}$ ensemble, a feedforward neural network (NN), and a gated recurrent unit neural network (GRU). The NN and GRU architectures are trained to minimize the one-step next-state prediction loss $\mathcal{L} =\mathbb{E}_{(s_t, s_{t+1}) \sim D} \left\| \hat{s}_{t+1} - s_{t+1} \right\|^2$, and serve as parametric baselines against which to compare the non-parametric $k\text{NN}$ models.

    \paragraph{\texorpdfstring{Evaluation terminology.}{Evaluation terminology.}}
    Because no simulator of the WTP is available, all evaluation is based on a temporally held-out slice of sensor data. For rollout evaluation, model-generated trajectories are compared with the recorded trajectory beginning from the same initial state, which we refer to as the \emph{true} trajectory. For hyperparameter selection, the held-out sequence is replayed in temporal order to emulate the data stream encountered during passive online deployment. We refer to this replay condition as the \emph{Online} setting and treat it as the ground-truth reference.

    We evaluate each calibration model according to its ability to (1) generate realistic long-horizon rollouts and (2) support hyperparameter selection for a TD(0) prediction agent \citep{sutton2018rlTextbook}:

    \begin{itemize}
        \item \textbf{Rollout Quality:} We generate 30k-step model rollouts and compare them qualitatively with the corresponding true trajectories. Pointwise error metrics can be difficult to interpret because otherwise similar trajectories may differ in phase, timing, or frequency. We therefore emphasize visual comparison in this initial experiment. Section~\ref{sec:RealWorldDeployment} additionally considers dynamic time warping (DTW), while recognizing that a single scalar distance cannot fully characterize long-horizon rollout quality.

        \item \textbf{Hyperparameter Selection:} We sweep Adam learning rates for a TD(0) prediction agent trained from scratch using either a calibration model or the Online data stream. We report NRMSE for the NN and GRU models and worst hyperparameter rank for the ensembled $k\text{NN}$ models. We then compare the resulting learning-rate sensitivity curves with the Online curves to assess how well each calibration model preserves hyperparameter rankings and sensitivity.
    \end{itemize}

    \subsection{Results}
    \label{sec:wtp_results}

\begin{figure}[htb]
        \centering
        \includegraphics[width=\textwidth]{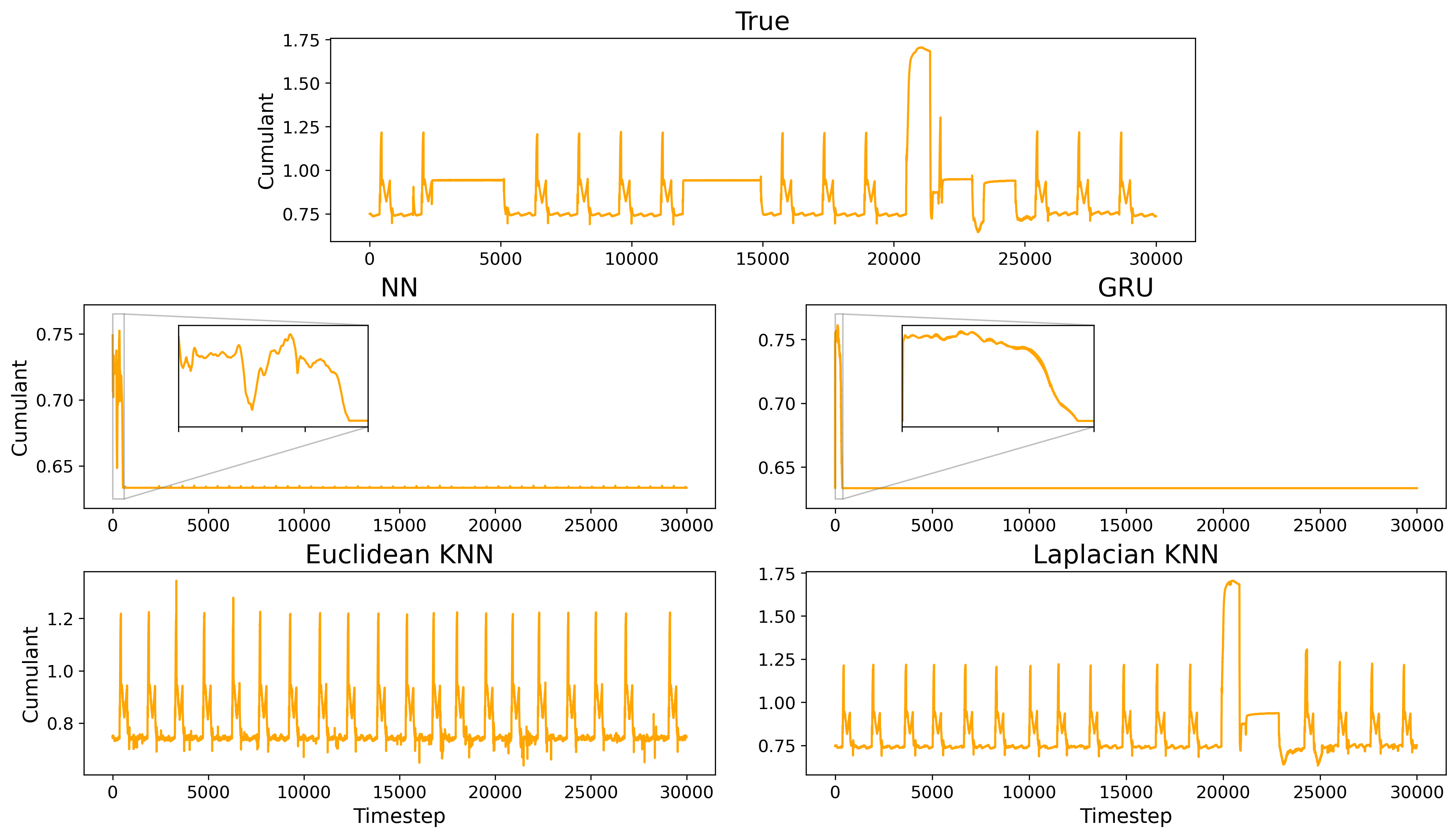}
        \caption{PIT300 sensor (membrane pressure) rollouts from the held-out test data (true) and calibration models. Each model is rolled out for 30k steps, beginning from the same start state.}
        \label{fig:wtpPIT300rollouts}
      \end{figure}

    In Figure~\ref{fig:wtpPIT300rollouts}, we compare the rollouts of the four calibration models to the true PIT300 trajectory from the held-out test data. We find that the $k\text{NN}$ models produce sensor trajectories that resemble the true data for PIT300, with the Laplacian $k\text{NN}$ showing closer alignment overall. For other sensors, rollouts are less accurate across all models, but the $k\text{NN}$ models still outperform NN and GRU baselines, which tend to collapse after $\sim$50--100 timesteps.

\begin{figure}[htb]
        \centering
        \includegraphics[width=\textwidth]{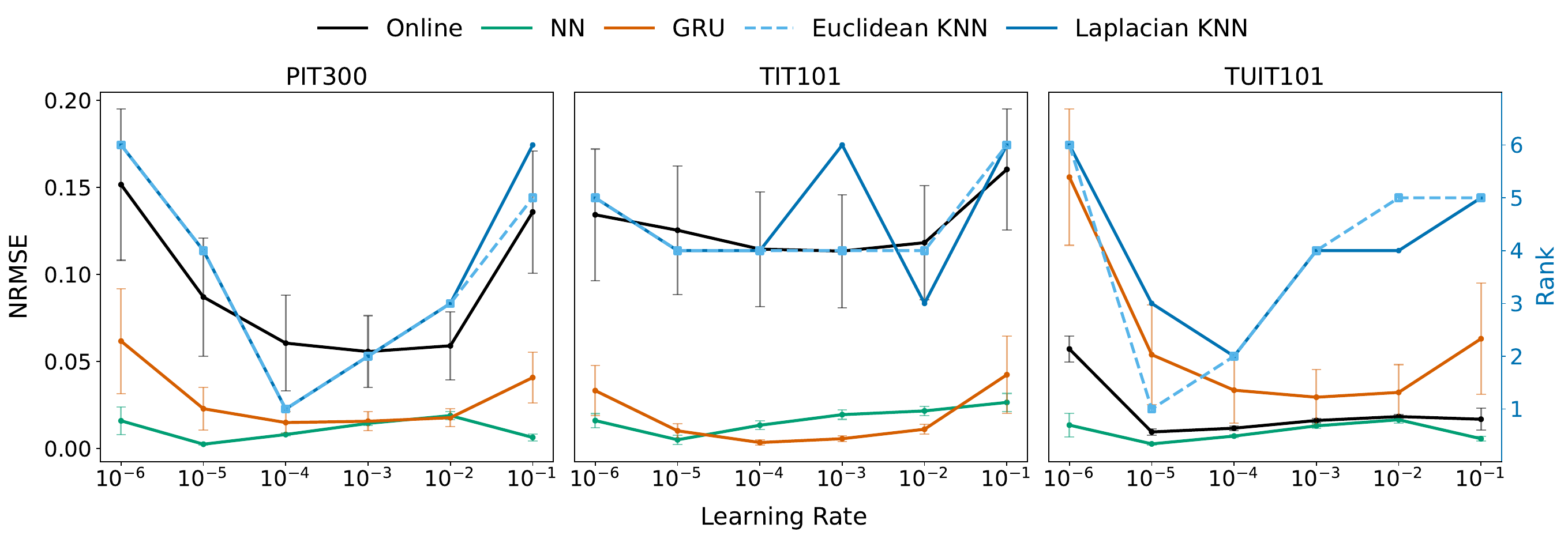}
        \caption{Learning rate sensitivity curves for a TD(0) prediction agent in the Online setting (replayed held-out data) and calibration models using the 1-week WTP dataset. Mean NRMSE and 95\% confidence intervals are shown for the non-ensembled models. For the ensembled $k\text{NN}$ models, worst rank is shown on the right y-axis. Because the two y-axes measure different quantities, curves should be compared by their shape and the relative ordering of learning rates, not by absolute values.}
        \label{fig:wtp1WeekSweeps}
      \end{figure}

    Figure~\ref{fig:wtp1WeekSweeps} shows learning rate sensitivity curves for each model. $k\text{NN}$ results are plotted on a separate rank axis in accordance with the ensembling strategy described in Section~\ref{subsec:knn_calibration_model} --- similarity in \emph{shape} of the sensitivity curves is what we use to determine agreement. Both $k\text{NN}$ models generally recover the correct hyperparameter rankings, with the Euclidean $k\text{NN}$ aligning most closely with the Online curves for TIT101 and TUIT101. The weaker performance of the Laplacian $k\text{NN}$ may stem from using a representation tuned for PIT300, highlighting the need for more robust Laplacian selection. As expected, the NN and GRU models show poor hyperparameter sensitivity, consistent with their low-quality rollouts.

%% file: sections/RealWorldDeployment.tex
\section{Towards Real-World Deployment}
\label{sec:RealWorldDeployment} 

Algorithms that perform well in controlled research experiments often face additional challenges in the complexity of real-world deployment. In this section, we take steps toward bridging this gap by exploring several modifications to the setting introduced in Section~\ref{sec:WaterTreatment}, broken down into three categories:

\paragraph{Scaling Up:}
In industrial settings like water treatment, years of offline sensor logs are available, presenting an opportunity to train models that capture non-stationarity, seasonality, and rare events. We extend the $k\text{NN}$ calibration model to a full year of WTP data, amounting to $\sim$32M samples at 1~Hz, which we sub-sample by a factor of 10 to yield a more manageable $\sim$3.2M samples. The model is built in two phases---KD-tree construction and neighbor table generation---resulting in a total complexity of $\mathcal{O}((d+k)n\log n)$ \citep{brown2014kdtree}, which takes approximately 10 hours on a 2 GHz Quad-Core Intel Core i5 processor for $d=142$, $k=3$, and $n=3.2$M. While computationally intensive, this process is a one-time cost, and can be further accelerated via dimensionality reduction (e.g., PCA or autoencoders), prototype selection~\citep{wilson2000reduction}, or approximate nearest-neighbor methods~\citep{indyk1998approximate, arya1998optimal}.

\paragraph{Agent Pre-training:}
In real-world settings where no simulator exists, we aim to maximize the utility of offline data. This typically involves pre-training an agent to avoid learning from scratch at deployment, after which the agent can continue to adapt online. We refer to this as the \emph{fine-tuning} setting, and shift our focus to selecting the fine-tuning learning rate. Since the same offline data is also used to construct a calibration model, we adopt a simple partitioning strategy to separate data for pre-training and calibration, reducing overlap and better simulating a realistic transfer scenario.

\paragraph{Distribution Shift:}
\label{par:DistributionShift}

\begin{figure}[htb]
    \centering
    \includegraphics[width=0.8\textwidth]{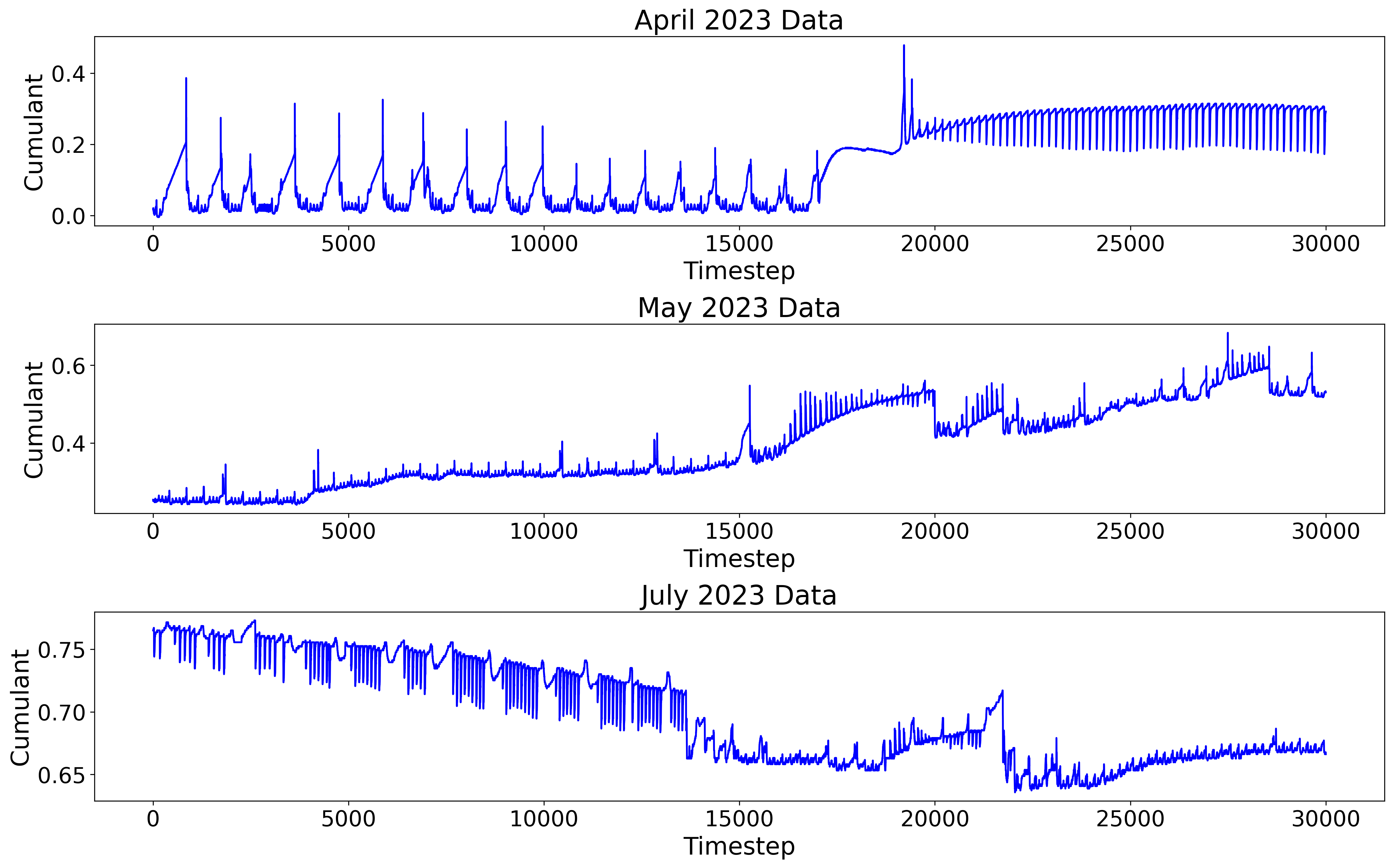}
    \caption{TIT101 sensor values across several time periods in the WTP dataset. Each plot shows roughly 3.5 days of data.}
    \label{fig:wtpTestMonths2023TIT101}
\end{figure}

System dynamics in real-world settings like the WTP can vary significantly over time due to factors such as rainfall, temperature, filter condition, and sensor drift. To evaluate how well calibration models handle such changes, we construct test sets that begin one week, one month, and three months after the training period, which correspond to April 2023, May 2023, and July 2023 respectively. As illustrated in Figure~\ref{fig:wtpTestMonths2023TIT101}, sensor patterns such as those from TIT101 change meaningfully across time, posing a challenge for generalization.

In general, it is not guaranteed that the calibration model will be able to simulate an arbitrary deployment period, since the distribution shift may be too large. However, we hypothesize that selecting a rollout start state that is representative of a deployment period can encourage the calibration model to simulate the dynamics of that period. We consider three possible strategies for selecting rollout start states $\mathcal{S}_0 = \{s_0^{[1]}, \dots, s_0^{[r]}\}$, where $r$ is the number of rollouts we perform:

\begin{enumerate}[label=(\roman*)]
    \item Randomly selecting $\mathcal{S}_0$ from the entire dataset (baseline method).
    \item Selecting $\mathcal{S}_0$ from the same calendar month in a previous year (e.g., July 2022 for a July 2023 deployment), if available.
    \item Selecting samples from the online deployment period and finding their nearest neighbours in the offline dataset.
\end{enumerate}
In our experiments, we focus on (iii) with (i) as a baseline, as (ii) assumes seasonal consistency that may not hold due to sensor drift or evolving plant conditions. Our use of model ensembling also means that each individual start state will not exist in all ensemble models, further complicating (ii). Note that (iii) is an oracle-like strategy: it assumes access to samples from the deployment period, which would not be available when selecting hyperparameters ahead of deployment. We adopt it here as a diagnostic tool, to test whether targeted start state selection can prompt the calibration model to simulate a specific distribution shift under best-case conditions. In practice, a natural approach would be to use the most recent plant readings prior to deployment as representative start states, since these are always available and likely closest in distribution to the upcoming deployment period.

\subsection{Experiments}

To put the preceding ideas into practice, we evaluate our Laplacian $k\text{NN}$ calibration model through two key questions:


\paragraph{Does scaling up the $\bm{k}\text{NN}$ calibration model improve generalization?}\label{par:ScalingUpExperiments}


We compare a $k\text{NN}$ model trained on one year of data (\textit{12-month $k\text{NN}$}), with one trained on the one-week dataset from Section~\ref{sec:WaterTreatment} (\textit{1-week $k\text{NN}$}). For each test period (April, May, and July 2023), we sample 30 random states and use their nearest neighbors under the Laplacian distance metric as rollout start states, following technique~(iii) from the Distribution Shift paragraph of Section~\ref{sec:RealWorldDeployment}. We evaluate rollouts both qualitatively (visual inspection) and quantitatively using DTW, where a lower DTW score indicates closer resemblance to the true sensor sequence from the test period.

\begin{figure}[!htb]
	\centering
	\includegraphics[width=\textwidth]{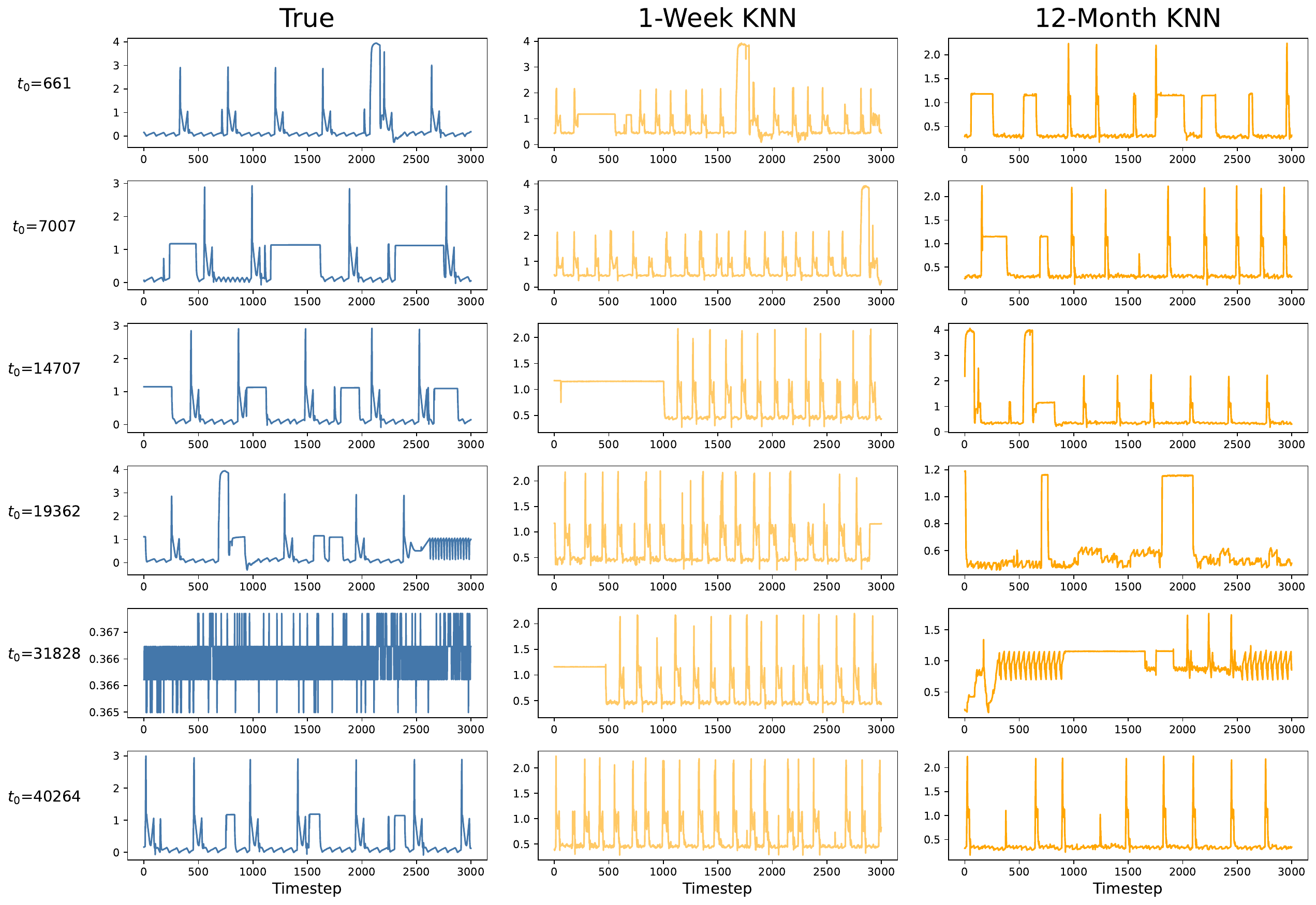}
	\caption{PIT300 rollouts for the 12-month and 1-week WTP $k\text{NN}$ calibration models using start states $t_0$ from the April 2023 dataset. True rollouts from those start states are shown in the leftmost plots.}
	\label{fig:wtp12mVs1weekKNNPIT300April2023}
\end{figure}

Figure~\ref{fig:wtp12mVs1weekKNNPIT300April2023} shows a subset of PIT300 rollouts using start states from the April 2023 test set. While it is challenging to qualitatively compare rollouts, we note that the 12-month $k\text{NN}$ is able to capture a broader range of dynamics than the 1-week $k\text{NN}$, with similar results for TIT101 shown in Figure~\ref{fig:wtp12mVs1weekKNNTIT101May2023}, though not for TUIT101 (Figure~\ref{fig:wtp12mVs1weekKNNTUIT101May2023}). Quantitative results using DTW (Table~\ref{table:wtp_dtw_results}) show mostly consistent rankings across step patterns, but mixed results across models: the 12-month $k\text{NN}$ performs best on TIT101, the 1-week $k\text{NN}$ on TUIT101, while results on PIT300 are mixed. Overall, these findings suggest that while the 12-month model may offer broader behavioral coverage, its generalization advantage is not conclusive. While DTW was preferable to mean squared or absolute error due to alignment issues, alternate metrics may be needed for clearer assessment \citep{coblin2024calibmodels}.


\begin{table}[!htb]
\centering
\scriptsize
\resizebox{\textwidth}{!}{%
\begin{tabular} {c|c|c|c|c|c|c}
    \hline
    & \multicolumn{2}{c|}{PIT300} & \multicolumn{2}{c|}{TIT101} & \multicolumn{2}{c}{TUIT101} \\ 
    \hline
    & 1-Week & 12-Month & 1-Week & 12-Month & 1-Week & 12-Month \\
    \hline
    Symmetric2 & $\bm{414.02 \pm 0.02}$ & $483.24 \pm 0.02$ & $910.20 \pm 0.05$ & $\bm{782.05 \pm 0.04}$ & $\bm{154.78 \pm 0.01}$ & $194.86 \pm 0.01$ \\
    Asymmetric & $\bm{224.0 \pm 0.02}$ & $283.38 \pm 0.03$ & $523.05 \pm 0.05$ & $\bm{431.87 \pm 0.04}$ & $\bm{100.92 \pm 0.01}$ & $153.81 \pm 0.02$ \\
    SymmetricP1 & $805.51 \pm 0.04$ & $\bm{782.79 \pm 0.04}$ & $1031.66 \pm 0.05$ & $\bm{898.61 \pm 0.04}$ & $\bm{199.55 \pm 0.01}$ & $268.24 \pm 0.01$ \\
    RabinerJuang & $\bm{339.87 \pm 0.03}$ & $367.87 \pm 0.04$ & $521.66 \pm 0.05$ & $\bm{447.82 \pm 0.04}$ & $\bm{100.43 \pm 0.01}$ & $154.09 \pm 0.02$ \\
    \hline
\end{tabular}}
\caption{Dynamic time warping distances computed between rollouts from $k\text{NN}$ models (1-Week and 12-Month) and true rollouts, averaged over three test sets with 30 rollouts each for each model and sensor combination. Values are reported as mean $\pm$ 95\% confidence interval. Smaller distance is better, and the best model for a specific sensor is presented in bold font. The leftmost column shows the step pattern used for the DTW algorithm.}
\label{table:wtp_dtw_results}
\end{table}

\paragraph{Can the $\bm{k}\text{NN}$ calibration model simulate distribution shifts to support fine-tuning learning rate selection?}\label{par:DistributionShiftExperiments}

To simulate a fine-tuning scenario, we pre-train a TD(0) prediction agent on the first six months of data and use the remaining six months to construct a $k\text{NN}$ calibration model. We then use this model to guide selection of the fine-tuning learning rate for the pre-trained agent, with start states selected using technique~(iii) (\emph{Month Start States}), and using random start states as described in technique~(i) (\emph{Year Start States}) for comparison.

\begin{figure}[htb]
    \centering
    \includegraphics[width=\textwidth]{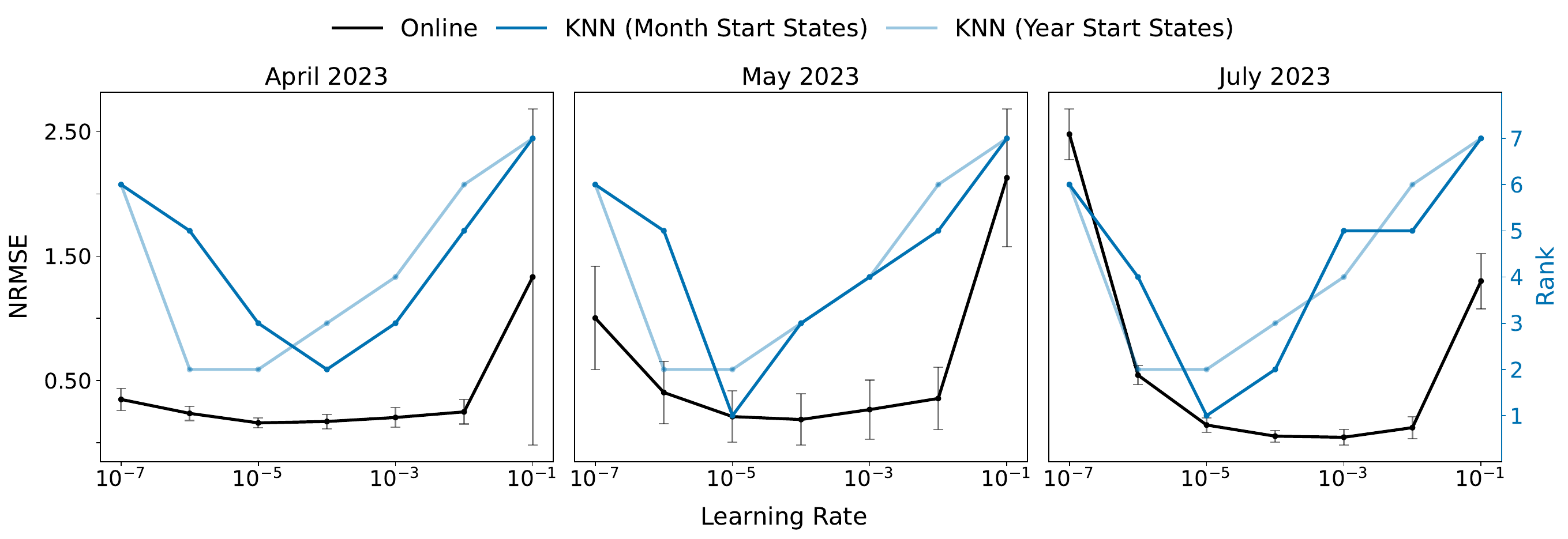}
	\caption{Fine-tuning learning rate sensitivity curves for the Online setting (replayed held-out data) and calibration models for PIT300 using six months for pre-training and six months for the calibration model. Mean NRMSE and 95\% confidence intervals are shown for the Online setting. For the ensembled $k\text{NN}$ models, worst rank is shown on the right y-axis. Because the two y-axes measure different quantities, curves should be compared by their shape and the relative ordering of learning rates, not by absolute values.}
	\label{fig:wtp6mSweeps}
  \end{figure}

The Online curves in Figure~\ref{fig:wtp6mSweeps} show that small learning rates yield good performance given a small distribution shift in the April 2023 test set. However, as the deployment period gets further away, the performance of the smallest learning rates deteriorates, indicating that the agent requires more adaptation. We find that the 6-month $k\text{NN}$ model is able to reflect a generic distribution shift, as shown by the learning rate curves giving best performance around \num{1e-4} to \num{1e-5}. However, it struggles to capture \emph{specific} shifts, as seen by its mismatch with the Online sensitivity curves. Additionally, performance between the month start states and full year start states is similar, suggesting limited benefit from our targeted start state rollout strategy. While these experiments offer a first step toward using calibration models under distribution shift, further work is needed to understand how to prompt the model to simulate dynamics from a specific deployment period, or even a period within its training data.

%% file: sections/Conclusion.tex
\section{Conclusion}
\label{sec:conclusion}

This work extends the calibration model framework for offline hyperparameter selection to a real-world industrial setting, focusing on sensor prediction tasks in a water treatment plant. We show that $k\text{NN}$-based calibration models can generate realistic long-horizon rollouts and preserve useful hyperparameter sensitivity. By investigating extensions involving large datasets, fine-tuning, and non-stationarity, we provide a proof of concept for leveraging offline data to guide online RL deployment.

However, important open questions remain. Under what conditions do extrapolating models fail? How should calibration models simulate distribution shifts associated with a target deployment period? To what extent is predictive accuracy necessary for preserving hyperparameter rankings? Are there more informative metrics than DTW for assessing long-horizon rollout quality? Addressing these questions will advance our understanding of calibration models and learned dynamics models more broadly, particularly in real-world systems.

%% file: sections/Supplementary.tex
\beginSupplementaryMaterials

\section{Laplacian Distance Metric}\label{sec:laplacian_distance_metric}

 Laplacian representations involve the graph Laplacian matrix $L$, which is defined as $L = D - A$, where $D$ is the degree matrix and $A$ is the adjacency matrix of the graph induced by the MDP, whose nodes correspond to state--action pairs. Laplacian representations have proved useful in RL for value-function approximation \citep{mahadevan2007proto}, option discovery \citep{machado2017laplacian}, and reward shaping \citep{wu2018laplacian}, among other applications.

Let $\mathbf{u}_1,\dots,\mathbf{u}_d$ be the first $d$ eigenvectors of the graph
Laplacian~$L$. Writing $x = (s, a)$ for a state--action pair, these define the representation

\begin{equation*}
  \psi : \mathcal{S} \times \mathcal{A} \longrightarrow \mathbb{R}^{d}, \qquad
  \psi(x) = \bigl[\,\mathbf{u}_1(x),\dots,\mathbf{u}_d(x)\bigr]^{\top},
\end{equation*}

where $\mathbf{u}_i(x)$ is the value of the $i$-th eigenvector evaluated at state--action pair $x$. The distance metric is then defined as

\begin{equation*}
    d(s_i, a_i, s_j, a_j) = \| \psi(x_i) - \psi(x_j) \|^2_2, \qquad x_i = (s_i, a_i),\; x_j = (s_j, a_j).
\end{equation*}

Analytically computing the eigenvectors of $L$ is typically not feasible for large graphs with an unknown transition function. Hence, we use an approximate method following the work in \citet{wu2018laplacian} and \citet{wang2022pesky}, which leverages spectral graph drawing to stochastically approximate the eigenfunctions of the Laplacian. Given a dataset $\mathcal{D}$, the graph drawing objective can be expressed as

\begin{equation*}
    \sum_{x_t \sim \mathcal{D}} \big\Vert \psi_{\theta}(x_t) - \psi_{\theta}(x_{t+1}) \big\Vert^2_2 + \sum_{x_i, x_j \sim \mathcal{D}} \Big((\psi_{\theta}(x_i)^T \psi_{\theta}(x_j))^2 - \big\Vert \psi_{\theta}(x_i) \big\Vert^2_2 - \big\Vert \psi_{\theta}(x_j) \big\Vert^2_2 \Big),
\end{equation*}

where $\psi_{\theta} : \mathcal{S} \times \mathcal{A} \rightarrow \mathbb{R}^d$ is the representation learned via a neural network with parameters $\theta$. Intuitively, this objective is composed of an \textit{attractive term} and a \textit{repulsive term}. The first term is attractive insofar as it encourages $\psi_{\theta}$ to map state--action pairs $x_t$ and their successors $x_{t+1}$ closely in the representation space --- this roughly captures temporal distance within an MDP. Conversely, the second term encourages independently sampled state--action pairs from the dataset to have orthogonal representations.

\section{Calibration Model Training Details}\label{supp:wtp_model_training}

\paragraph{$\bm{k}\text{NN}$ and Laplacian Models}

For each $k\text{NN}$ variant, we use a five-member LOBO ensemble with $k=3$. To preserve sequential dependencies, the dataset is deterministically partitioned into five disjoint, contiguous blocks. Each ensemble member is trained on all data except its corresponding held-out block. The full training and calibration process proceeds in three stages:

\begin{enumerate}
	\item \textit{(Laplacian variant only)} Train a neural network to map raw states into a dynamics-aware Laplacian representation.
	\item For each ensemble model, construct a KD-tree over its data subset in the learned representation space to support efficient nearest-neighbor queries.
	\item Precompute fixed neighbor tables from the KD-trees, enabling constant-time next-state predictions during rollouts.
\end{enumerate}

The underlying $k\text{NN}$ model follows \citet{wang2022pesky}; further implementation details for the LOBO variant in the context of the WTP are provided by \citet{coblin2024calibmodels}. Note that the LOBO ensembling method is referred to as bootstrapping in that work.

Candidate Laplacian representations are evaluated using two validation metrics. \textit{Dynamics awareness} measures how well the representation preserves temporal relationships between states, while \textit{representation uniqueness} detects degenerate representations in which distinct states collapse to similar embeddings. Among representations with a uniqueness score greater than 0.95, we select the one with the highest dynamics-awareness score. The corresponding hyperparameters are reported in Table~\ref{table:LaplacianHypers}.

\paragraph{NN Calibration Models}  
The NN calibration models used a two-layer feedforward neural network architecture with a shared set of hyperparameters across sensors. Models were trained on a one-step-ahead prediction objective, mapping the current state (and optionally action) to the next state. A grid search over learning rates, hidden sizes, and batch sizes was conducted, and the best hyperparameters were selected based on performance on a validation set --- see Table~\ref{table:FNNHyperparams}. Although multi-step prediction targets were explored, one-step predictions yielded better rollout performance in practice.

\paragraph{GRU Calibration Models}  
The GRU calibration models employed a two-layer recurrent architecture designed to capture temporal dependencies in the sensor data. Models were trained on one-step-ahead prediction using sequences of historical states, with a burn-in period and sequence length tuned for each dataset. Dropout and other regularization techniques were tested to improve long-horizon stability, but simple GRU models with tuned hidden sizes and sequence lengths produced the best results. As with the NN models, hyperparameters were selected via grid search, focusing on validation performance --- see Table~\ref{table:RNNHyperparams}.








\begin{table}[!htb]
	\centering
	\scriptsize
	\begin{tabular} {|c|c|c|c|}
	  \hline
	  \textbf{Hyperparameter} & \textbf{Symbol} & \textbf{Water 1-Week (Online)} & \textbf{Water 12-Month (Pre-training)} \\
	  \hline
	  Optimizer & - & Adam & Adam \\
	  Learning Rate & $\alpha$ & - & \num{1e-5} \\
	  Discount Factor & $\gamma$ & 0.99 & 0.9 \\
	  Batch Size & $B$ & 256 & 256 \\
	  Hidden Layers & - & 2 & 2 \\
	  Hidden Units & - & 256 & 512 \\
	  Replay Buffer Size & - & 1M & 3M \\
	  Train/Validation Split & - & - & 0.9/0.1 \\
	  Epochs & - & - & 1000 \\
	  \hline
	\end{tabular}
	\caption{Hyperparameters used for the TD(0) prediction agent in WTP experiments.}
	\label{table:TD0Hyperparams}
\end{table}

\begin{table}[!htb]
	\centering
	\scriptsize
	\begin{tabular} {|c|c|c|}
	  \hline
	  \textbf{Hyperparameter} & \textbf{Symbol} & \textbf{Water 1-Week} \\
	  \hline
	  Optimizer & - & Adam \\
	  Learning Rate & $\alpha$ & \num{1e-3} \\
	  Batch Size & $B$ & 256 \\
	  Hidden Layers & - & 2 \\
	  State Model Hidden Size & - & 512 \\
	  Epochs & - & 100 \\
	  \hline
	\end{tabular}
	\caption{NN calibration model training hyperparameters for the WTP.}
	\label{table:FNNHyperparams}
\end{table}

\begin{table}[!htb]
	\centering
	\scriptsize
	\begin{tabular} {|c|c|c|}
	  \hline
	  \textbf{Hyperparameter} & \textbf{Symbol} & \textbf{Water 1-Week} \\
	  \hline
	  Optimizer & - & Adam \\
	  Learning Rate & $\alpha$ & \num{1e-3} \\
	  Batch Size & $B$ & 256 \\
	  Hidden Layers & - & 2 \\
	  State Model Hidden Size & - & 512 \\
	  Burn-in Length & - & 0 \\
	  Sequence Length & - & 20 \\
	  State Model Epochs & - & 100 \\
	  \hline
	\end{tabular}
	\caption{GRU calibration model training hyperparameters for the WTP.}
	\label{table:RNNHyperparams}
\end{table}

\begin{table}[!htb]
	\centering
	\scriptsize
	\begin{tabular} {|c|c|c|c|}
	  \hline
	  \textbf{Hyperparameter} & \textbf{Symbol} & \textbf{Water 1-Week} & \textbf{Water 12-Month} \\
	  \hline
	  Optimizer & - & Adam & Adam \\
	  Learning Rate & $\alpha$ & \num{3e-4} & \num{1e-5} \\
	  Batch Size & $B$ & 256 & 256 \\
	  Hidden Layers & - & 2 & 2 \\
	  Hidden Units & - & 256 & 256 \\
	  Output Dimension & - & 64 & 64 \\
	  Training Steps & - & 100,000 & 200,000 \\
	  Train/Validation Split & - & 0.8/0.2 & 0.8/0.2 \\
	  Sequence Length & - & 20 & 20 \\
	  Kappa & $\kappa$ & 0.95 & 0.95 \\
	  Beta & $\beta$ & 5 & 5 \\
	  Zeta & $\zeta$ & 0.05 & 0.05 \\
	  \hline
	\end{tabular}
	\caption{Laplacian representation training hyperparameters for the WTP.}
	\label{table:LaplacianHypers}
\end{table}


\section{Experiment Details for Scaled Up Generalization Capabilities}

We compare a $k\text{NN}$ model trained on one year of data between March 31, 2022 and March 31, 2023 (\textit{12-month $k\text{NN}$}), with one trained on the one-week dataset from Section~\ref{sec:WaterTreatment} (\textit{1-week $k\text{NN}$}). For each test period (April, May, and July 2023), we sample 30 random states and use their nearest neighbors under the Laplacian distance metric as rollout start states, following technique~(iii) from the Distribution Shift paragraph of Section~\ref{sec:RealWorldDeployment}. Each rollout spans 3k steps. Hyperparameters are provided in Table~\ref{table:LaplacianHypers}.

\section{Experiment Details for Fine-Tuning Learning Rate Selection}

To simulate a fine-tuning scenario, we pre-train a TD(0) prediction agent on the first six months of data and use the remaining six months to construct a $k\text{NN}$ calibration model (i.e. an equal partitioning strategy). We then use this model to guide selection of the fine-tuning learning rate. Hyperparameters for the TD(0) prediction agent are provided in Table~\ref{table:TD0Hyperparams}.

For each learning rate, we perform 30 runs in the Online setting and 10 per ensemble model for the calibration model, resulting in 50 total runs per $\alpha$ in the ensemble setting. We report NRMSE averaged over the final 25\% of each run, so that early training error does not dominate the metric.

\clearpage
\section{Additional Results}\label{supp:wtp_calibration_model_performance}

\begin{figure}[htb]
    \centering
    \includegraphics[width=\textwidth]{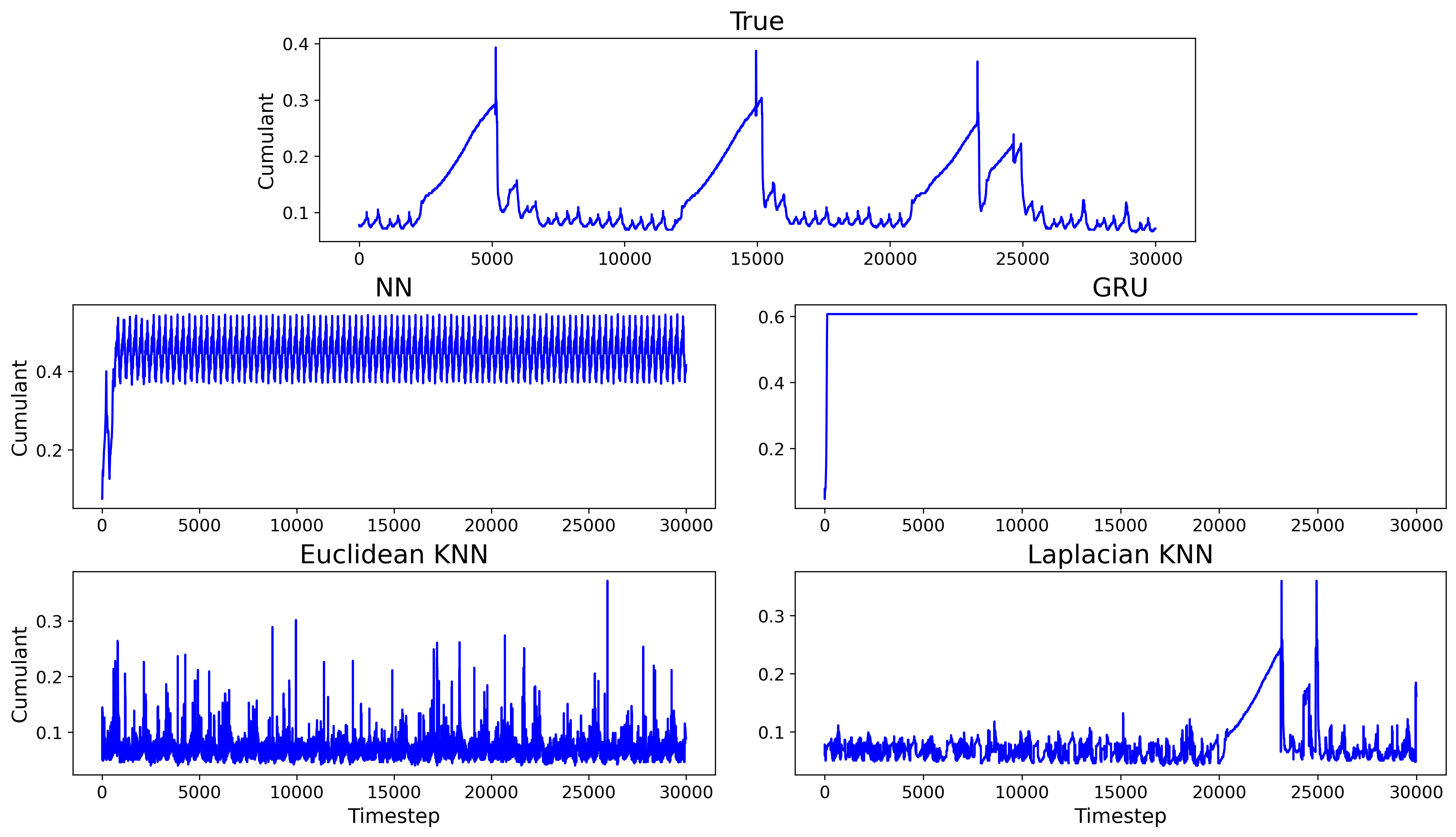}
    \caption{TIT101 sensor (influent temperature) rollouts from the held-out test data (true) and calibration models. Each model is rolled out for 30k steps, beginning from the same start state.}
    \label{fig:wtpTIT101rollouts}
\end{figure}

\begin{figure}[htb]
    \centering
    \includegraphics[width=\textwidth]{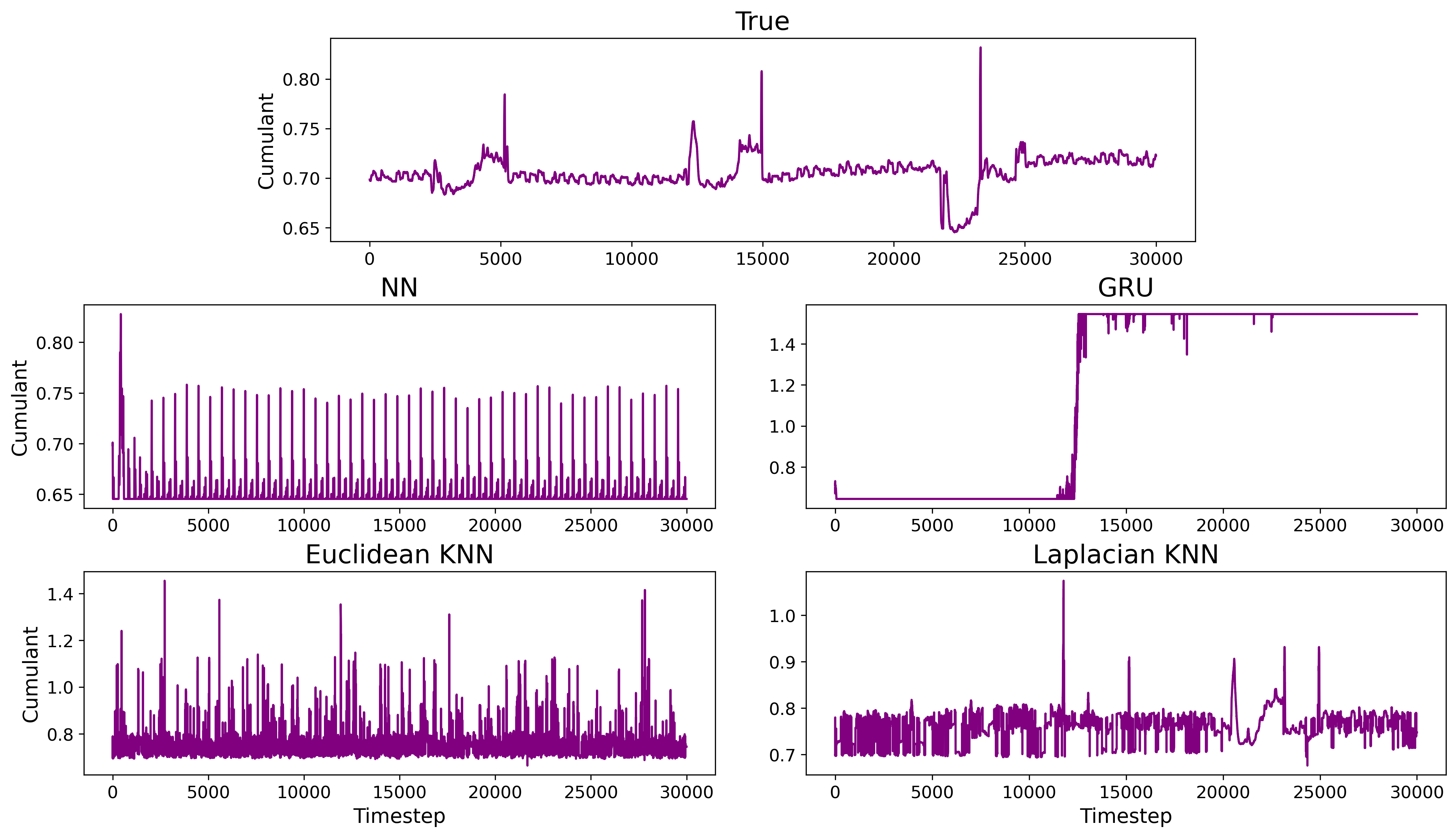}
    \caption{TUIT101 sensor (influent turbidity) rollouts from the held-out test data (true) and calibration models. Each model is rolled out for 30k steps, beginning from the same start state.}
    \label{fig:wtpTUIT101rollouts}
\end{figure}

\begin{figure}[!htb]
	\centering
	\includegraphics[width=\textwidth]{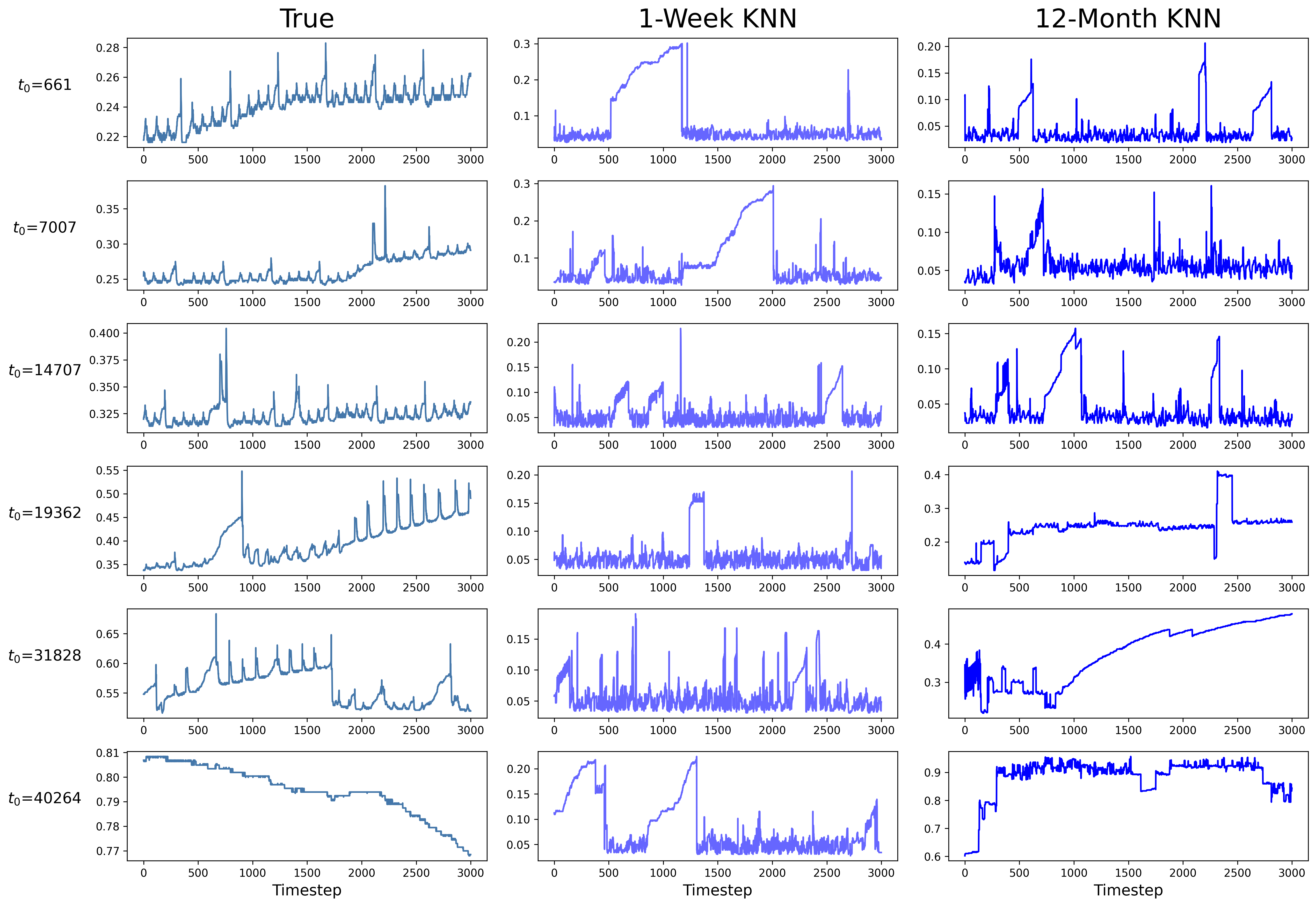}
	\caption{TIT101 rollouts for the 12-month and 1-week WTP $k\text{NN}$ calibration models using start states $t_0$ from the May 2023 dataset. True rollouts from those start states are shown in the leftmost plots.}
	\label{fig:wtp12mVs1weekKNNTIT101May2023}
\end{figure}

\begin{figure}[!htb]
	\centering
	\includegraphics[width=\textwidth]{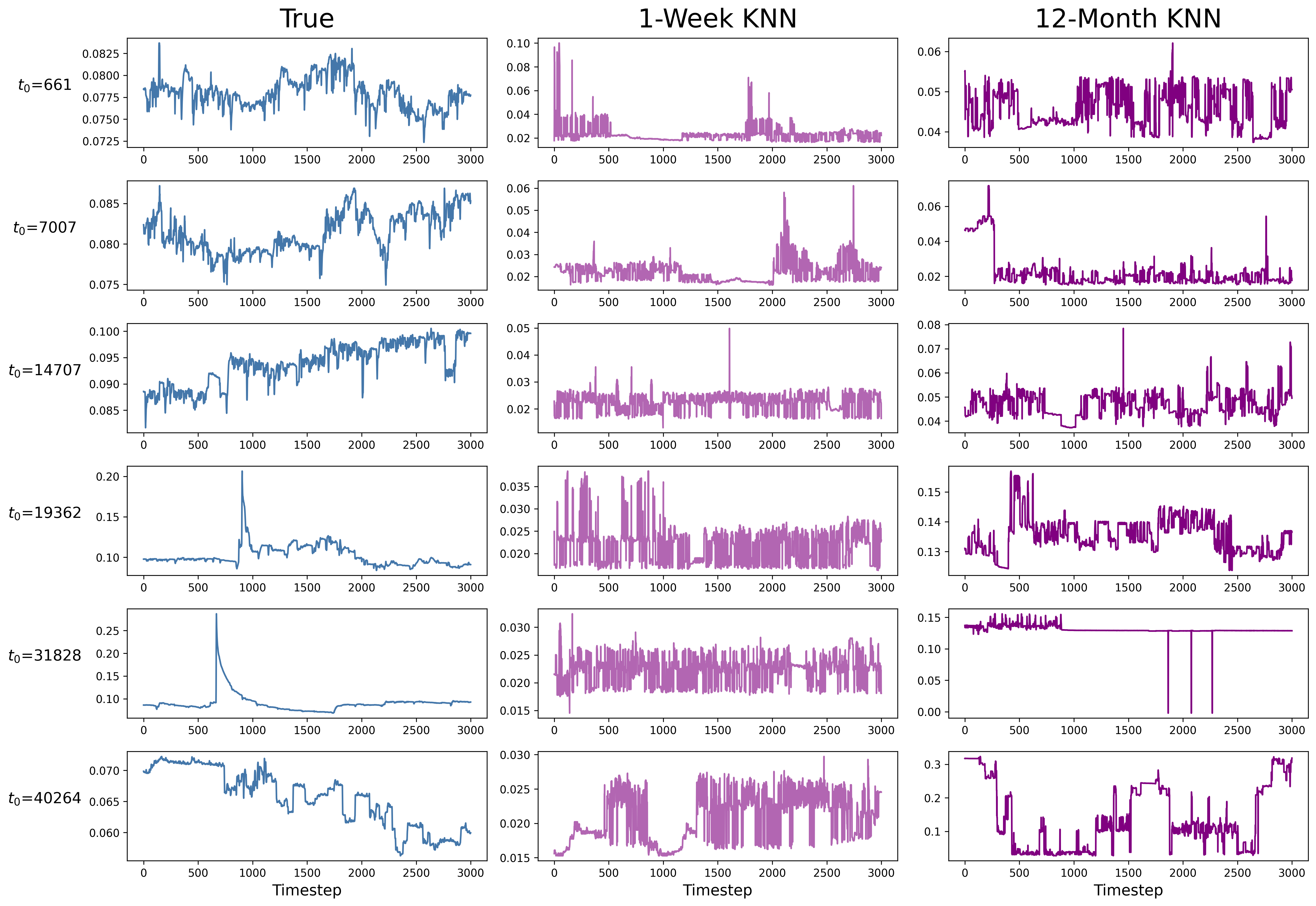}
	\caption{TUIT101 rollouts for the 12-month and 1-week WTP $k\text{NN}$ calibration models using start states $t_0$ from the May 2023 dataset. True rollouts from those start states are shown in the leftmost plots.}
	\label{fig:wtp12mVs1weekKNNTUIT101May2023}
\end{figure}
